# SOME ASPECTS OF TESTING PROCESS FOR TRANSPORT STREAMS IN DIGITAL VIDEO BROADCASTING

Radu ARSINTE (*)     Ciprian ILIOAEI (**)

*(\*) Technical University Cluj-Napoca , Tel: +40-264-595699, Str. Baritiu 26-28, Radu.Arsinte@com.utcluj.ro*
*(\*\*) Tedelco SRL Cluj-Napoca , Calea Turzii 42, Tel:+40-264-595613 , ciprian.ilioaei@tedelco.ro*

**Abstract:** This paper presents some aspects related to the DVB (Digital Video Broadcasting) investigation. The basic aspects of DVB are presented, with an emphasis on DVB-T version of standard. The main purpose of this research is to analyze the way that the transmission of the transport streams is realized in case of the Terrestrial Digital Video Broadcasting (DVB-T). To accomplish this, first, Digital Video Broadcasting standard is presented, and then the main aspects of DVB testing and analysis of the transport streams are investigated. The paper presents also the results obtained using two programs designed for DVB analysis: Mosalina and TSA.

## I. DIGITAL VIDEO BROADCASTING

The current generation of TV signal broadcast standards is based on digital data compression and digital data transmission. This provides both higher image quality and better bandwidth utilization than classic analog color TV broadcasting standards such as PAL, NTSC, and SECAM.

In January 1995, the Digital Video Broadcasting (DVB) project organized by the European Broadcasting Union (EBU) has published a set of formal standards, which define the new Digital Video Broadcast system. These DVB standards are the technical basis for implementing digital TV transmission in Europe, Asia, Australia, and many others regions of the world. DVB is proved to be the best candidate for a single global digital TV broadcasting standard. DVB-T broadcasting started in some countries of Europe (Germany, UK) and is considered for adopting in other (including Romania).

Future updates of the standards have improved the specifications, but the main lines remained the same.

Current DVB standards describe digital TV transmission over satellite and cable and modulation standards for terrestrial broadcast. DVB standards cover the system design and the modem standards for high bandwidth video data transmission as well as several auxiliary functions like Teletext, electronic program guides, and conditional access. The compression technique used by the DVB system is the ISO MPEG-2 algorithm.

The block diagram of a DVB receiver (generally called "set-top box") is presented in figure 1. The block schematic is almost the same for terrestrial, satellite or cable receivers, the main difference being in the RF stages of the diagram.

There are three basic standards of DVB:
- DVB-C is designated for TV and data transmission via cable
- DVB-S is used for transmission via satellite
- DVB-T is used for terrestrial TV transmission

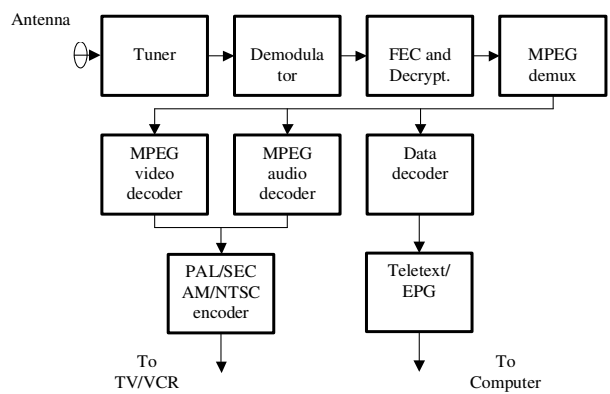

*Fig. 1. Block schematic of DVB receiver*

**DVB-S**

DVB-S([1],[2],[4]) is a satellite-based delivery system designed to operate within a range of transponder bandwidths (26 to 72 MHz) accommodated by European satellites such as the Astra series, Eutelsat series, Hispasat, Telecom series, Tele-X, Thor, TDF-1 and 2, and DFS [3]. DVB-S is a single carrier system, with the payload (the most important data) at its core. Surrounding this core are a series of layers intended not only to make the signal less sensitive to errors, but also to arrange the payload in a form suitable for broadcasting. The video, audio, and other data are inserted into fixed-length MPEG transport-stream packets. This packetized data constitutes the payload. A number of processing stages follow:
• The data is formed into a regular structure by inverting synchronization bytes every eighth packet header.
• The contents are randomized.
• Reed-Solomon forward error correction (FEC) overhead is

added to the packet data. This efficient system, which adds less than 12 percent overhead to the signal, is known as the outer code. All delivery systems have a common outer code.
• Convolutional interleaving is applied to the packet contents.
• Another error-correction system, which uses a punctured convolutional code, is added. This second error-correction system, the inner code, can be adjusted (in the amount of overhead) to suit the needs of the service provider.
• The signal modulates the satellite broadcast carrier using quadrature phase-shift keying (QPSK).

In essence, between the multiplexing and the physical transmission, the system is tailored to the specific channel properties. The system is arranged to adapt to the error characteristics of the channel. Burst errors are randomized, and two layers of forward error correction are added. The second level (inner code) can be adjusted to suit the operational circumstances (power, dish size, bit rate available, and other parameters).

**DVB-C**

The cable network system, known as DVB-C, has the same core properties as the satellite system, but the modulation is based on quadrature amplitude modulation (QAM) rather than QPSK, and no inner-code forward error correction is used [3]. The system is centered on 64-QAM, but lower-level systems, such as 16-QAM and 32-QAM, also can be used. In each case, the data capacity of the system is traded against robustness of the data.

Higher-level systems, such as 128-QAM and 256-QAM, also may become possible, but their use will depend on the capacity of the cable network to cope with the reduced decoding margin. In terms of capacity, an 8 MHz channel can accommodate a payload capacity of 38.5 Mbits/s if 64-QAM is used, without spillover into adjacent channels.

**DVB-T**

DVB-T [3] must fulfill the requirements working in the following conditions:
- very variable signal-to-noise ratio
- large-scale multi-path effects (reflections from nearby
- house walls, etc. attenuate certain frequencies)
- very overcrowded frequency spectrum, interference from nearby analog TV channels and sometimes remote stations on the same frequency band.

The selected DVB-T modulation scheme has the following characteristics:
- COFDM (Coded Orthogonal Frequency Division Multiplex).
- In this technique, a fast Fourier transform is used to generate a broadcast signal with thousands of mutually orthogonal QAM modulated carriers. A single symbol carries therefore several kilobits of information. A guard interval between symbols allows echos to pass by before the receiver starts the detection of the next symbol.
- 8192 or 2048 carrier frequencies, each modulated using a QAM with between 4 and 64 Symbols
- 8 MHz bandwidth
- outer FEC: Reed Solomon FEC RS(204, 188, T=8)
- outer convolutional interleaving
- inner FEC: Convolutional Code (1/2, ..., 7/8)
- inner interleaving

With thousands of separate carrier frequencies, the typical elimination of some frequencies due to multi-way path reflections in terrestrial transmission becomes tolerable as the interleaving and FEC spreads the payload information uniformly across the entire bandwidth. OFDM modulation is in particular also well suited for mobile and indoor reception, as well as the operation of single-frequency networks (SFN). In a single frequency network, many neigbouring transmitters over a large area broadcast the exact same signal carefully synchronized. The resilience of the modulation scheme against echoes can handle this as merely an extreme multi-path propagation as well, resulting in much better use of the frequency spectrum as safety distances between transmitters operating on the same frequency become less critical.

**II. TRANSPORT STREAM STRUCTURE**

General diagram of the encoding process for DVB-T is presented in figure2.

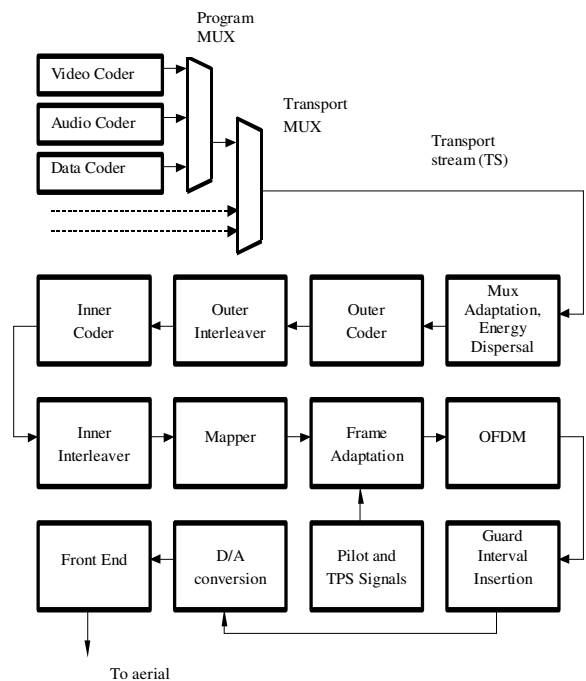

*Fig.2. DVB-T coding*

In this general schematic our interest was focused on TS (transport stream structure) for a closer look. Transport stream is generated from multiple PES (Packetized Elementary Streams). Decoding this structure brings information on data content, error rates and a lot of other information useful in quality link evaluation.

References ([1]…[5]) describes in detail this complex structure, so in the following rows only the basic aspects of TS are presented.

Figure 3 presents the structure of MPEG-2 and Transport packets.

The transport stream is composed from time-multiplexed packets from different streams. The MPEG-2 packet is a 188-byte packet, consisting of a 4-byte header leaving 184 bytes for payload, as shown in figure 3a. The small packet length is more suited for high error mediums as the errors thus affects less data, although the header represents a higher overhead for such a packet size. The header, as shown in figure 3, contains a Sync byte used for random access to the stream. It also contains a Program ID (PID), which allows for the identification of all packets belonging to the same data stream, or alternatively it provides a mean for multiplexing data streams within transport streams. It may be viewed as the equivalent of the port number field in UDP packets. Finally, the Continuity Counter field (CC) may be viewed as the equivalent of the RTP sequence number. It is incremented by one for each packet belonging to the same PID therefore allowing for the detection of missing packets.

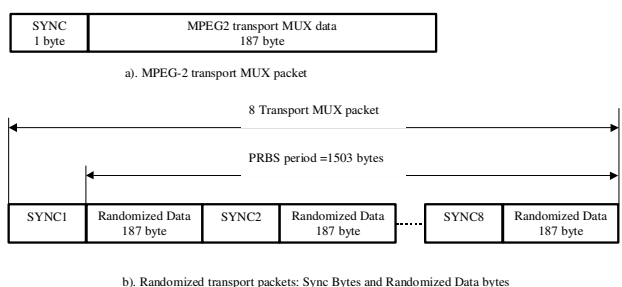

*Fig.3. Packet structure in DVB*

## III. SHORT DESCRIPTION OF THE TEST PROGRAMS

In this section we'll describe the Transport Stream Analyser (TSA). This software product allows to the user to analyse DVB transport stream recordings on simple PC equipment.

The TSA presents the data using a hierarchical structure, making it easy to locate the information you wish to examine, and is useful both for analysis and as an educational tool.

The TSA software was tested in this research on Windows 98 platform but it works on Windows NT and Windows 2000.

The software was designed using the concept of plugins. These are software components that add a feature to the TSA, for example the decoding of subtitles is handled by a single plugin. The format of the plugin interface is publicly available, so that third parties will be able to extend the functionality of the TSA.

The current release of the TSA software, allows the following tests:
• Transport Packet
• Video
• Monitor

The results of testing are presented in section IV of the paper.

The second program analyzed, Mosalina, is more complex. It allows to the user to perform also picture quality measurements on the incoming stream, and not only data structure analysis.

Briefly it is described in the following rows.

Mosalina is a complete solution for quality assessment and technical analysis of MPEG2 elementary streams. The main components are presented.

### Toolbox
The Mosalina toolbox is the central hub to all actions. The icons on the tool bars launch common processes with a single click. User friendly dialogs then guide the user through any additional steps that are necessary.

### Project Manager
Project Manager collates and maintains all files required by Mosalina and monitors them for validity. Project Manager also collects a reference of any output from the Asset Scanner and Batch Processor components. Demo mode will disable all save functions.

### Stream Analyzer
Stream Analyzer is the most powerful utility within the Mosalina package. Stream Analyzer uses metadata created by the software (MIF data, Thumbnail Images) and Thresholds from the selected bitstream to analyze and browse a stream. It also contains playback facilities for viewing streams. Demo mode is limited to 400 frames.

### MIF Compiler
The MIF Compiler is used to generate the MIF (MPEG2 Information File) and Thumbnail image files that are used by the Stream Analyzer and Report Generator components. Demo mode is limited to 400 frames.

### Report Generator
Report Generator is used to create reports about a bitstream. Selected Thresholds can be parsed simultaneously to generate a comprehensive report detailing pictures, GOPs or seconds that lie outside defined tolerances. Report Generator will also generate quality certificates if a stream successfully passes all user defined thresholds. Demo mode is limited to 400 frames.

### Stream Player
The MPEG-2 video decoder provides a solution for software decoding of any MPEG-2 file, at any resolution (SD/HD), in any MPEG profile/level, in real-time (depending on bitrate and PC speed). Multiple players can be used simultaneously.

### SI Voyager
SI Voyager provides the user with a visual hierarchical description of the PSI/SI tables contained within a transport stream. Compliancy problems with the MPEG system or DVB specifications are highlighted. Available in demo mode, but only by using the demo file provided.

### Asset Scanner
The Asset Scanner component is used to control Asset Locators. Each locator can search a specified path and option subdirectories for any assets that Mosalina can process. The frequency of the scanning can be set so that assets created at run-time can be automatically handled by the system without any user interaction required. Scanned assets can be optionally asent to the Batch Processor.

### Batch Processor
The Batch Processor is a component designed to ease the user task multiple assets for processing without further need for interactions. The results from Batch Processor will automatically be added to Project Manager if it is open.

### Thumbnail Preview
The Thumbnail Preview window provides a visual overview of the stream. This window can be used to easily check whether stream material overlaps.

### Program Stream Demux
The Program Stream demultiplexer will search a given stream and present the video PID's found to the user. The user can then select the desired PID/s (all PID's will be demultiplexed in batch mode) to disk. Filenames will automatically be suggested and can be overridden by the user after the PID search. Not available in demo mode.

**Transport Stream Demux**

The Transport Stream demultiplexer will search a given stream and present the video PID's found to the user. The user can then select the desired PID/s (all PID's will be demultiplexed in batch mode) to disk. Filenames will automatically be suggested and can be overridden by the user after the PID search. Demo mode will only allow demultiplexing of the transport stream demo file provided.

### IV. RESULTS

For the transport stream analysis, as discussed, two different programs have been used: Mosalina and TSA (Transport Stream Analyzer)[5]. The results obtained after the transport stream decoding are the same, but Mosalina has more options compared to TSA.

Next, we'll show the results for specific test transport stream, named "flux2.l2t". This file can be found on most DVB dedicated sites. Notice that the transport packets in this case are delivered for two programs.

**Table 1** – transport packets for the program 0X1241(4673)

| Index | PID Value | Packet type | Error indicator | Transport priority | Index | PCR | Continuity counter |
|---|---|---|---|---|---|---|---|
| 0 | 520 | video | 0 | 0 | 3 | 0 | 0 |
| 1 | 730 | audio | 0 | 0 | 1 | - | 0 |
| 2 | 731 | audio | 0 | 0 | 1 | - | 0 |
| 3 | 732 | audio | 0 | 0 | 1 | - | 0 |
| 800 | 800 | private | 0 | 0 | 1 | - | 0 |

**Table 2** – transport packets for the program 0X1181(4481)

| Index | PID value | Packet type | Error indicator | Transport priority | Adaptation control | PCR | Continuity counter |
|---|---|---|---|---|---|---|---|
| 0 | 520 | video | 0 | 0 | 3 | 0 | 0 |
| 1 | 730 | audio | 0 | 0 | 1 | - | 0 |
| 2 | 731 | audio | 0 | 0 | 1 | - | 0 |
| 3 | 732 | audio | 0 | 0 | 1 | - | 0 |
| 800 | 800 | private | 0 | 0 | 1 | - | 0 |

**Table 3** – PSI Sections of the PSI tables

| Index | PID value | PSI type | Section length | Version number | Section number | Table ID |
|---|---|---|---|---|---|---|
| 4 | 0 | PAT | 21 | 0 | 0 | 0X00 |
| 485 | 16 | NIT | 163 | 0 | 0 | 0X40 |
| 647 | 17 | SDT | 121 | 21 | 0 | 0X42 |
| 808 | 18 | ELT | 282 | 3 | 0 | 0X4e |
| 1616 | 20 | TDT | 9 | 9 | 48 | 0X70 |
| 162 | 268 | PMT | 81 | 0 | 0 | 0X02 |
| 324 | 269 | PMT | 81 | 9 | 0 | 0X02 |

Notice that for this particular transport stream we have received 14 different packets:

- one video packet
- 3 audio packets
- 7 signaling packets
- 3 additional packets

**Fields specifications**

- *PID value*: is assigned to each packet and it's different from one transport stream to another.
- *Packet type*: depends on the information type-video, audio, data or signaling.
- *Error indicator* : indicates if any kind of error exists in the transport stream transmission .
- *Transport priority*: indicates the importance of the information transmitted within the transport stream (0-not very important).
- *PCR*: Program Clock Reference
- *Continuity counter*: used for the indication of a possible interruption in data transmission.
- *Section length*: usually on 12 bytes, indicates the bytes number of the section used for the PSI tables transmission.
- *Version number*: on 5 bytes, indicates which version of the section has been used.
- *Section number*: specifies the section number used for the transmission of signaling tables.
- *Table ID*: is different for every signaling table type(ex: for PMT is set on 0x02,for NIT is set on 0x0010).

### V. CONCLUSIONS

1.The transport packets from one transport stream can be delivered to more than one program.

2. For one particular transport stream, after the de-multiplexing process a video packet is obtained, several audio packets and also additional packets like PSI, reserved etc.

3. The PID value is very important because this way we can say to which transport stream belongs one received packet.

**Short comparison between TSA and Mosalina**
- They both perform analysis of one transport stream, indicating the transport packets type, that are received in Online or Offline mode;
- TSA has a much more common interface, is very simple and has less options than Mosalina. One specific option is the live monitoring of the packets received;
- Mosalina has a structure based on the components like: Toolbox, Report Generator etc. TSA instead has a tree structure view with two options: Programs and Packets;
- Working with Mosalina, among the analyze of the received packets, it is possible to find other DVB transmission parameters like: bit rate, PAR value, QSC value etc.;
- With both programs we can work Online or Offline. TSA has the Softel Transport Stream Capture Utility facility which allows the capture of one transport stream in real time;
- If we reefer strictly on the transport packets received, both programs indicate the same packets for whom we can also find out the PES (Packetzed Elementary Stream) information or PSI Sections if necessary;

**Advantages and disadvantages of using TSA**
- TSA has less options than Mosalina;
- With TSA we can't identify the DVB transmission parameters so we can't appreciate its quality;
- Even if TSA hasn't so many options it is much simpler, the time for one analysis is very short and the information about a transport packet is almost complete;

A major advantage can be the fact that the TSA structures the received packets after their PID value.

**Future work**
- Extending the test pattern in decoded image (MPEG-2 quality assessment)
- Building drivers for several DVB receivers to realize a real time analysis.
- Extending the results in DVB-S and DVB-C with minor modifications